\documentclass[conference]{IEEEtran}
\IEEEoverridecommandlockouts
\usepackage{cite}
\usepackage{multirow}
\usepackage{amsmath,amssymb,amsfonts}
\usepackage{algorithm}
\usepackage{algpseudocode}
\usepackage{graphicx}
\usepackage{textcomp}
\usepackage{xcolor}
\usepackage{mwe}
\usepackage{subfig}
\def\BibTeX{{\rm B\kern-.05em{\sc i\kern-.025em b}\kern-.08em
    T\kern-.1667em\lower.7ex\hbox{E}\kern-.125emX}}
\makeatletter
\let\old@ps@headings\ps@headings
\let\old@ps@IEEEtitlepagestyle\ps@IEEEtitlepagestyle
\def\confheader#1{%
\def\ps@IEEEtitlepagestyle{
\old@ps@IEEEtitlepagestyle
\def\@oddhead{\strut\hfill#1\hfill\strut}
\def\@evenhead{\strut\hfill#1\hfill\strut}
}
\ps@headings
}
\makeatother
\confheader{
\small{Proceedings of the 9th RSI International Conference on Robotics and Mechatronics (ICRoM 2021), Nov. 17-19, 2021, Tehran, Iran} 
}
\usepackage[pscoord]{eso-pic}
\newcommand{\placetextbox}[3]{
\setbox0=\hbox{#3}
\AddToShipoutPictureFG*{ \put(\LenToUnit{#1\paperwidth},\LenToUnit{#2\paperheight}){\vtop{{\null}\makebox[0pt][c]{#3}}}
}
}
\placetextbox{.23}{0.055}{\textbf{\small{978-1-6654-2094-5/21/\$31.00~\copyright 2021 IEEE}}}

\begin{document}

\title{A Gaussian Process-Based
Ground Segmentation for Sloped Terrains}

\author{\IEEEauthorblockN{ Pouria Mehrabi}
\IEEEauthorblockA{\textit{Advance Robotics \& Automated Systems
(ARAS)}\\
K. N. Toosi University of Technology, Tehran, Iran.\\
p.mehrabi@email.kntu.ac.ir}\and \IEEEauthorblockN{Hamid D. Taghirad}
\IEEEauthorblockA{\textit{Advance Robotics \& Automated
Systems (ARAS)}\\
K. N. Toosi University of Technology, Tehran, Iran. \\
taghirad@kntu.ac.ir}}
\maketitle
\begin{abstract}
A Gaussian Process ($\mathcal{GP}$) based ground segmentation method
is proposed in this paper which is fully developed in a
probabilistic framework. The proposed method tends to obtain a
continuous realistic model of the ground. The LiDAR
three-dimensional point cloud data is used as the sole source of the
input data. The physical realities of the data are taken into
account to properly classify sloped ground as well as the flat ones.
Furthermore, unlike conventional ground segmentation methods, no
height or distance constraints or limitations are required for the algorithm to be applied to take all the regarding physical behavior of the ground into account.
Furthermore, a density-like parameter is defined to handle
ground-like obstacle points in the ground candidate set. The
non-stationary covariance kernel function is used for the Gaussian
Process, by which Bayesian inference is applied using the maximum
\textit{A Posteriori} criterion. The log-marginal likelihood
function is assumed to be a multi-task objective function, to
represent a whole-frame unbiased view of the ground at each frame.
Simulation results show the effectiveness of the proposed method
even in an uneven, rough scene which outperforms similar Gaussian
process-based ground segmentation methods.
\end{abstract}


\section{Introduction}
The imminent advent of level five autonomy of driver--less cars is
not favorable anymore to be considered as one's far-fetched
ambition. Autonomous Land Vehicle's (ALV) may provide many
opportunities from empowering the ability of remote exploration and
navigation in an unknown environment to establishing driver--less
cars that may navigate autonomously in urban areas while being safer
by compensating human driving faults. Therefore, ALVs are expected
to conquer the urban transportation industry soon enough if they
meet certain performance prerequisites. Safe maneuvers in urban
areas through different obstacles and during different possible
scenarios would be considered as one of such highly consequential
prerequisites. To establish such a driver--less car, developed
methods shall be reliable and real--time implementable. This
reliability is obtained only if ALVs are capable of obtaining a
detailed and reliable perception of the environment they are tending
to perform in.
Detailed perception of the urban environment in which driver-less
cars are operating is often obtained by pre-and post-processing of
raw data frames coming from its sensors. Different objects and
obstacles in each scene must be detected and classified to construct
the needed perception. While different detection and classification
methods are proposed to tackle this problem, ground segmentation
remains a vital part of this classification procedure as the vehicle
will plan its future possible routes and reactions on a possible
space that is defined based on the ground. Thus, ALVs shall
efficiently recognize the ground in unknown environments they tend
to operate. Although a reliable ground segmentation procedure shall
be applicable in environments with both flat and sloped terrains the
issue remains not fully solved in the literature.

Gaussian process regressions are useful tools for the implementation
of Bayesian inference which relies on correlation models of inputs
and observation data~\cite{C.E.Rasmussen2006}. Although light detection
and ranging (LiDAR) sensors are
commonly used in ALVs, data resulting from these sensors do not
inherit smoothness, and therefore, stationary covariance functions
may not be used to implement Gaussian regression tasks on these
data. Different methods for segmentation have been proposed in the
literature with these constraints~\cite{Zermas2017, Shin2017,
Korchev2016, Asvadi2016, Chen2014, Himmelsbach2010, Moosmann2009}.

In \cite{Zermas2017}, the ground surface is obtained in an iterative
routine, using deterministic assignment of the seed points. In
\cite{Shin2017}, the ground segmentation step is put aside to
establish a faster segmentation based on Gaussian process
regression. Thus, A 2D occupancy grid is used to determine surrounding
ground heights, and a set of non-ground candidate
points are generated. Reference \cite{Korchev2016} handles real-time
segmentation problem by differentiating the minimum and maximum
height map in both rectangular and a polar grid map. In
\cite{Asvadi2016} a geometric ground estimation is obtained by a
piece-wise plane fitting method capable of estimating arbitrary
ground surfaces. 

In \cite{Chen2014} a Gaussian process-based methodology is used to perform ground estimation by segmenting the
data with a fast segmentation method firstly introduced by
\cite{Himmelsbach2010}. The non-stationary covariance function from
\cite{Paciorek2003} is used to model the ground observations while
no specific physical motivated method is given for choosing
length scales. Paper~\cite{Chu2019} proposed a human-centric ground segmentation method
based on processing the point cloud in both vertical and horizontal directions.
Paper \cite{Moosmann2009} proposes a fast
segmentation method based on local convexity criterion in non-flat
urban environments. The jump-convolution process is utilized in~\cite{Shen2021} to obtain
a fast estimation of ground surface using LiDAR data without any consideration of sloped ground segments.

These methods are either estimating ground piece-wise and with the
local viewpoint or by labeling all the individual points with some
predefined criterion. Except \cite{Chen2014} non of the above-mentioned methods, gives a continuous model for the predicted ground.
Furthermore, none of them gives an exact
method to extract local characteristics of non-smooth data. However,
efficient ground segmentation has to be done considering the physical
properties of the data including non-smoothness of the LiDAR data
and ground condition in every data frame. Thus, even direct
assignment of slope values to length scales in \cite{Chen2014} does
not lead to the proper classification of ground points coming from
sloped areas. This is because the careful specification of covariance structures is critical
especially in non-parametric regression tasks~\cite{Stein2005}.

LiDAR data consists of three-dimensional range data which is
collected by a rotating sensor, strapped down to a moving car. This
moving sensor causes non-smoothness in its measured data, which may
not be taken into account using common stationary covariance
functions. Although in some methods such as \cite{Chen2014} a
Gaussian process-based method for ground segmentation with
non-stationary covariance functions are proposed, adjusting
covariance kernels to accommodate the physical property of the
ground segmentation problem needs further investigation.

Length-scales may be defined as the extent of the area that data
points may affect on each other \cite{Plagemann2008}. Length-scale
values play a significant role in the quality of the interpretation
that the covariance kernel gives about the data. A constant length
scale may not be used with LiDAR data due to the non-smoothness of
the collected point cloud. Different methods are proposed to adjust
length scales locally for non-stationary covariance functions by
assuming an exact functional relationship for length-scale values
\cite{Zhang2017, Fuglstad2013, Fuglstad2015}. The ground
segmentation method proposed by \cite{Chen2014} assumes the length
scales to be a defined function of line features in different
segments. This is not sufficient because no physical background is
considered for the selection of functional relationships and this
function might change and fail to describe the underlying data in
different locations.
\begin{figure}[!t]
    \centering
    \includegraphics[scale=0.3]{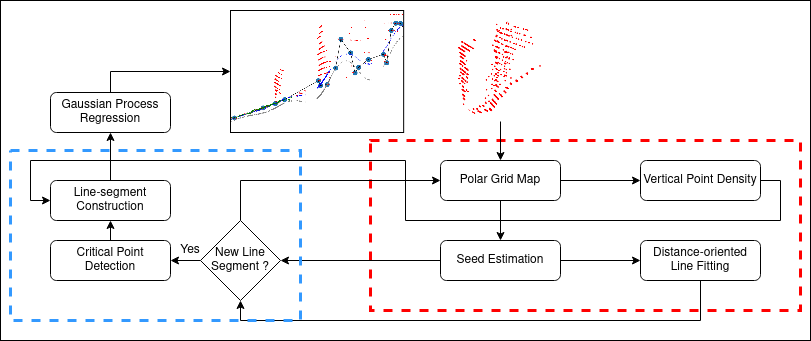}
    \caption{Block diagram of the proposed ground segmentation method.}
    \label{figure:diagram}
\end{figure}

The quality of the ground is
not elaborated in each segment with disregard for adjacent segments.
Furthermore, The physical properties of the data are taken into
account for the method to be able to properly classify sloped ground
as well as the flat ground. An intuitive guess of length-scale
values are represented by assuming the existence of hypothetical
surfaces in the environment that every bunch of data points has
resulted from measurements of these surfaces. Thus, unlike other
ground segmentation methods in the literature, no height or distance
constraints or limitations are applied to the data which enables the
data to take all the available information into account. This
enables the method to gain a more clear and realistic view of the
ground as it insists on the fact that the ground might not always be a flat segment or restricted from having heights more than certain values.

The proposed method tends to obviate the difficulties of ground segmentation in sloped areas by considering the two-dimensional line fitting method as the core of the line extraction algorithm in which the exact distance of the points to the line is taken into account instead of vertical distance which is the prevailing method in this field. Furthermore, a physically motivated line extraction algorithm is introduced which is intuitively compatible
with what happens in real-world urban scenarios. 

For the ground points placed adjacent to or beneath the obstacle points in 2D segments, a density-like parameter is introduced as a parameter in the GP kernel to make the ground estimation more reliable for these areas. This parameter gives a higher density-like value to the points related to line-segments with an obstacle in their proximity.

The three-dimensional LiDAR data is segmented using a radial grid
map. Each LiDAR scan is divided into $m$ segments. Then
each segment is divided into $n$ bins. Then, points
with the lowest height in each bin are considered to be the ground
candidate points in each segment.  The proposed method
is tested on KITTI \cite{Geiger2013} data set and the data set
provided by \cite{Lai2009} which is labeled. The method is shown to
outperform similar Gaussian process-based ground segmentation
methods and at the same time gives a proper estimation of sloped
ground. All modules and algorithms are implemented and developed
from scratch using C++ using Qt Creator. Furthermore, 3D point
clouds are handled by the use of Point Cloud Library (PCL). A
Gaussian Process Regression ($\mathcal{GP}$) library is also
developed in C++ based on the math provided in
\cite{C.E.Rasmussen2006}.

\section{Radial Grid Map} In order to arrange all the points at
each segment according to their radial distance to the origin of
$xy$--plane, each segment is divided into different bins with
respect to its radial range~\cite{Himmelsbach2010}. Each particular
segment will be divided unevenly into $N$ bins because the minimum
and maximum of radial distances are different between individual
segments and $n_{th}$ bin in $m_{th}$ segment $b_n^m$ covers the
range from $r_n^{min}$ to $r_n^{max}$. The set $P_{b_n^m}$ is
constructed which contains all of the three-dimensional points that
are mapped into the $n_{th}$ bin of the $m_{th}$ segment
\cite{Himmelsbach2010, Chen2014}. Furthermore the set
$P_{b_n^m}^\prime$ containing the corresponding two-dimensional
points is constructed by,
\begin{equation}\label{equation:four}
P_{b_n^m}^{\prime} = \{ p_i^{\prime} = (r_i,z_i)^T \; |\; p_i \in
P_{b_n^m}\},
\end{equation}
where, $r_i = \sqrt{x_i^2+y_i^2}$ is the radial range of
corresponding points. The \textit{ground candidate set} $PG_m$, being the first intuitive guess for
obtaining initial ground model is constructed by collecting the
point with the lowest height at each bin as the ground candidate in
that bin \cite{Chen2014}:
\begin{equation}
PG_m = \{p_i^\prime \; | \; p_i^\prime \in P_{b_n^m}^{\prime}, z_i =
\min(\mathcal{Z}_n^m)\},
\end{equation}
where, $\mathcal{Z}_n^m$ denotes the set of height values of all
points in the corresponding bin and segment. Furthermore, in each
bin a vertical segmentation is applied. Each bin is divided into $j$
vertical segments spread from minimum height $z_{min}^{m,n}$ to
$z_{max}^{m,n}$. Then the number of points in each of these vertical
segments are calculated and averaged on the range to obtain
$\tilde{\rho}_n^m$.
\begin{equation}\label{equation:fourPrime}
\tilde{\rho}_n^m = {\sum_{i=1}^{j}(\rho_i)}/j 
\end{equation}
Where $\rho_i$ is obtained by dividing the number of points in each vertical segment by the area of that segment.

\section{Gaussian Process Regression}
Gaussian processes ($\mathcal{GP}$) are stochastic processes with
any finite number of their random variables being jointly Gaussian
distributed. In this paper, Gaussian process regression is being
used as a tractable tool to put prior distributions over a nonlinear
function that relates the ground model to the radial location of
points. Gaussian processes are thoroughly defined by their mean and
covariance function:
\begin{equation}
f(x) \sim \mathcal{GP}\big( m(x),k(x,x^\prime)\big)
\end{equation}

The predictive distribution of Gaussian process nonlinear regression
is obtained from the joint distribution of the measurement and the function
values \cite{C.E.Rasmussen2006}, in which,
\begin{eqnarray}
&\bar{f_*} = K_{\mathcal{X}_*}\big(K+\sigma_n^2I\big)^{-1}y, \nonumber \\
&\textrm{cov} = K^* - K_{\mathcal{X}_*}\big(K +
\sigma_n^2I\big)^{-1}K_{\mathcal{X}_*}^T.
\end{eqnarray}
Log marginal likelihood or  evidence $P(y|\mathcal{X})$ is obtained
under Gaussian process assumption that the prior is Gaussian
$f|\mathcal{X}\sim \mathcal{N}(0,K)$:
\begin{equation}
\log P(f|\mathcal{X}) =
-\frac{1}{2}f^TK^{-1}f-\frac{1}{2}\log|K|-\frac{n}{2}\log(2\pi)
\end{equation}

\section{Physically Motivated Ground Segmentation}
Different kinds of terrains might be wended by an ALV, in different
kinds of environments and applications. Thus, flat and gradient
roads and grounds might be faced by an ALV in urban scenarios. In
this paper, a Gaussian process regression-based method is developed
to consider both flat and sloped terrains in urban road driving
scenes. This method estimates ground for separated segments with a
multi-task objective function to take the whole-frame condition of
ground into account.

In Gaussian process regression methods \textit{covariance kernels}
define how the parameters are related to each other and how they are
being affected by each other due to their specific mathematical
model at different extents. Although simple covariance kernels are
powerful tools for Bayesian inference, they fail to consider
local-smoothness, since in their simplest version they assume the
data to be stationary. On the other hand, LiDAR data inherits
"input-dependent smoothness", meaning that its data does not bear
smooth variation at every part and direction of the environment,
thus the stationarity assumption fails to describe this data
precisely. Therefore, covariance functions with constant
length-scales are not suitable for LiDAR point cloud since for
example flat grounds must have a length-scale that is more widely
valid than a rough ground or the points coming from different parts
of the data may not be assumed to behave regarding the same covariance
kernel. The non-stationary covariance function originally
represented by \cite{Paciorek2003} is used to model ground in this
paper.
\begin{eqnarray}\label{equation:nonscov}
&K(r_i,r_j) = \sigma_f^2.\big[\mathcal{L}_i^2\big]^{\frac{1}{4}}
\big[\mathcal{L}_j^2\big]^{\frac{1}{4}}
\big[ \frac{\mathcal{L}_i^2 +
 \mathcal{L}_j^2 }{2}\big]^\frac{-1}{2}\nonumber \\
& \times  \exp\bigg(\frac{-(r_i-r_j)^2}{[  \mathcal{L}_i^2 +
\mathcal{L}_j^2]}\bigg),
\end{eqnarray}

Local characteristics of this covariance function is calculated using
local line features by assuming length-scales to be of the form
$\mathcal{L}_i=ad_i$, where $a$ is assumed to be one of the
hyper-parameters of the regression problem.
\begin{eqnarray}\label{d_i}
&d_i = \frac{1}{\tilde{\rho}_n^m} \log\bigg(\frac{1}{g_i}\bigg)
\end{eqnarray}

\subsection{Physically-Motivated Line Extraction}
Different line extraction algorithms are being utilized in different
robotics applications with some of them being more generally
accepted and utilized. Although these algorithms are widely used,
some applications need to use different versions of
them~\cite{Gao2018}, \cite{Chen2014}, \cite{Siadat1997}. Line
extraction algorithms are previously used in different ground
segmentation methods to enable the distinction of the ground and
obstacle points. Thus, although these methods are effective in segmenting near-flat ground points, they fail to properly
recognize ground points coming from sloped ground sections or
gradient roads. \cite{Chen2014} states that despite using a
non-stationary covariance function as kernel, their method does not
work in the existence of sloped terrains. This can be due to the usage of the Incremental line
extraction algorithm which is elaborated in \cite{Nguyen2005} and
utilized in \cite{Chen2014} to estimate ground surface. The incremental line extraction algorithm lacks the
efficiency needed for the detection of sloped terrains. Thus a physically
motivated line extraction algorithm along with a two-dimensional line fitting method is introduced which is
intuitively compatible with what happens in real-world urban
scenarios. The proposed line extraction algorithm relies on the fact
that in urban structures, the successive lines of each laser scan
should be considered independently. For example, if some structures
are found in the data that shows a $3^{\circ}$ slope for a range of
radial distance ($2m$) and the algorithm is decided that the last
point of this series is a "critical point", the parameters of the
next line will start to construct from scratch and without
dependency to previous segments. This is because the ground
candidate points are successive points coming from different bins.
Therefore, a sudden change of structure is more important than the
overall behavior of some cluster of points as they may be related to
a starting point of an obstacle or sloped ground.

\paragraph{Definition of Critical Points}
To overcome the problem of sloped terrain detection in the ground
segmentation task, the proposed line extraction algorithm operates
based on finding some \textit{\textbf{critical points}} among the
ground candidate points set in each segment $PG_m$. The critical
points are defined to be the points at which the behavior of the
successive ground candidate points change in a way that can be
flagged as \textit{unusual}. This unusual behavior happens in the
areas that the ground meets the obstacle or as well as the areas at
which the road starts to elevate during a slope or gradient section
of its. These critical points, therefore, partition each segment
into different sections between each two successive critical points.
The 2D points between two successive critical points form a
\textbf{\textit{line-segment}}. These line-segments should be chosen
carefully as they play a significant role in the line extraction
algorithm and the further interpretation of the ground. On the other
hand, while large sloped line-segments relate to non-ground
structures, the low sloped ones may relate to the ground. Therefore,
The conditions listed below must be met for a ground point to be
considered as a critical point:
\begin{itemize}
    \item \textit{The slope of fitted line}: the slope of the fitted line for the potentially ground-related line segments must be greater than a threshold $\zeta_b$.
    \item \textit{Distance from point to the fitted line:} distance
    from points of each line-segment to the fitted line must be less
    than a given threshold $\delta$.
    \item \textit{Horizontal Distance of the points:} The horizontal distance of each two successive points must not exceed a given threshold $\zeta_m$. This is set to prevent including breakpoints.
    This threshold is set concerning each segment and about the radial size of each segment.
    \item \textit{Smoothness of $\tilde{\rho_n^m}$}: The average number of vertical points must not have a sharp change during each line-segment.
\end{itemize}
\paragraph{Distance--oriented Line Regression}
The distance-oriented line regression method is utilized as the core
line fitting algorithm. The standard least square method breaks down
when the slope of the line is almost vertical or when the slope of
the line is large but finite~\cite{Stein1983}. The least-square method assumes that only the dependent value is subject to error
thus the distance of the points to the estimated value $(\hat{y} -
y)^T(\hat{y} - y)$ is utilized. This assumption makes the algorithm
very sensitive to the position of the independent value especially
in larger slopes as depicted in Figure~\ref{fig_prependicular}. In
these cases, a small inaccuracy in the value of the regressor will
lead to greater uncertainty in the value of the regressed parameter
if the least square is used as the line fitting algorithm.

On the other hand, in many applications such as ground segmentation,
this assumption fails to be true as both coordinates are subject to
errors. Thus, the least square method is insufficient for sloped
terrain ground point detection as the method tends to obtain critical points in
areas with larger slopes and the accuracy of detection are of high
importance here. The distance-oriented line regression method takes
the exact distance of the points to the line into account. The orthogonal least square algorithm is utilized in~\cite{Gao2018} as
an alternative for the least square method. The method assumes that
both parameters have the same error while this assumption is not
valid for line-segment extraction as the 2D value is derived by
implementing manipulation on original 3D data. Furthermore, the
Non-stationarity assumption of the method denies any similarity of
errors in both directions. The mathematical model $\eta = \alpha +
\beta \xi$ is assumed to describe the linear relationship of two
underlying variables. If both of the variables are observed subject
to a random error, the relation of these measurements are as
follows:
\begin{figure}[!t]
\centering
\includegraphics[scale=0.5]{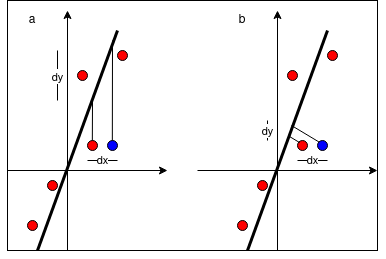}
\caption{The least square method have larger errors in sloped
regions. A same small error in horizontal displacement, results
greater error in \textit{y} value in \textbf{\textit{a}} (with least
square method) than in \textit{\textbf{b}} (with proposed fit line
method.}. \label{fig_prependicular}
\end{figure}
\begin{eqnarray}
&r_i = \xi_i + \epsilon_{r_i} \\
&z_i = \eta_i + \epsilon_{z_i} = \alpha + \beta \xi_i + \epsilon_{z_i}
\end{eqnarray}
In which $\epsilon_{r_i}\sim\mathcal{N}(0, \sigma^2)$ and
$\epsilon_{z_i}\sim\mathcal{N}(0, \lambda\sigma^2)$. The maximum
likelihood estimate for $\alpha$ and $\beta$ is derived. The
"log-likelihood" has the following form:
\begin{eqnarray}
&&L = -\frac{n}{2}log(4\pi^2)-\frac{n}{2}log(\lambda\sigma^4)\nonumber \\
&&-\frac{\Sigma_{i=1}^n(r_i-\xi_i)^2}{\sigma^2}-\frac{\Sigma_{i=1}^n(z_i-\alpha
-\beta\xi_i)^2}{2\lambda\sigma^2}
\end{eqnarray}
Differentiating the log-likelihood function with respect to
${\alpha, \beta, \xi, \sigma}$ and solving for $\frac{\partial
L}{\partial \star} = 0$ yields the numerical estimate of the line parameters.

\begin{center}
\begin{algorithm}
  \scriptsize
\begin{algorithmic}[1]
\State \textbf{INPUT} $PG_m$
\State \textbf{OUTPUT:} $g_i^m$, $criticalP$
\State $l \gets size(PG_m)$
\State $n_s \gets 0$
again:
 \For { $i = n_s ; to ; l - 1 $ }
        \State $j \gets 1$
        \State $P \gets Seed(i,j)$
        \State $P^\prime \gets P + Seed(i+j,1)$
        \State $(a, b, b) \gets line(P)$
        \State $(a^\prime, b^\prime, c^\prime) \gets line(P^\prime)$
        \If {$isCritical( (a, b, b), (a^\prime, b^\prime, c^\prime))$}
        \State $criticalP \gets line(ns , i)$
        \State $g_i\gets criticalP.slope$
        \State $n_s \gets i + j + 1$
        \State goto again;
        \EndIf
        \EndFor
    \end{algorithmic}
            \label{algo:line}
                 \end{algorithm}
\end{center}

\subsection{Problem Formulation}
Nonlinear Gaussian process regression problem is to recover a
functional dependency of the form $y_i=f(x_i)+\epsilon_i$ from $n$
observed data points of the training set $\mathcal{D}$. The set
$PG_m \{ (r_i,z_i)\}$ contains all of two-dimensional ground
candidate points in the segment $m$ which the final ground model is
inferred based on. The predictive ground model
$P_m(z_*|r_*,\mathcal{R}_m)$ for segment $m$ is built using the
ground candidate set as the input for the Gaussian process
regression task.

The covariance function \ref{equation:nonscov} is utilized as the
$\mathcal{GP}$'s kernel. The length-scale parameters $\mathcal{L}$
is set considering the evaluated line-segments. This enables the
predicted continuous ground model to show a reasonable distance to
the minimum points set in non-ground sections. The predictive
distribution of the height $z_*$ at the arbitrary test input
location $r_*$ is obtained using the predictive ground model. Then
the predicted ground values are compared to test input from the raw
data to label ground points.

\subsection{Problem Formulation}
\paragraph{Physically--Motivated Length--Scales}
Length-scales for each point in the ground candidate  set as well as
in the raw data are obtained as follows:
\begin{eqnarray}\label{equation:length-scales}
&\mathcal{L_i} =
     \begin{cases}
       \frac{\alpha}{\tilde{\rho}_n^m} \log\bigg(\frac{1}{g_i}\bigg) &\quad \text{if } g_i\ge g_d\\
       \alpha \log\bigg(\frac{1}{g_d}\bigg) & O.W. \\
     \end{cases}
\end{eqnarray}
The parameter $\alpha$  is the hyper-parameter for the regression
task which will be learned based on the data. The $g_d$ threshold is
set to prevent very large length-scale values when the slope is
small and near zero. The term $\frac{1}{\tilde{\rho}_n^m}$ is
related to the density of the points above each segment-line. This
term adds an extra penalty-like value to the length scales. This
makes sure that a reasonable distance will be kept with ground
candidate points in high-density areas.
\begin{figure}[t]
\begin{minipage}{.5\linewidth}
\centering
\subfloat[]{\label{main:a}\includegraphics[trim={1cm 0 1.5cm 1cm},clip,scale=.275]{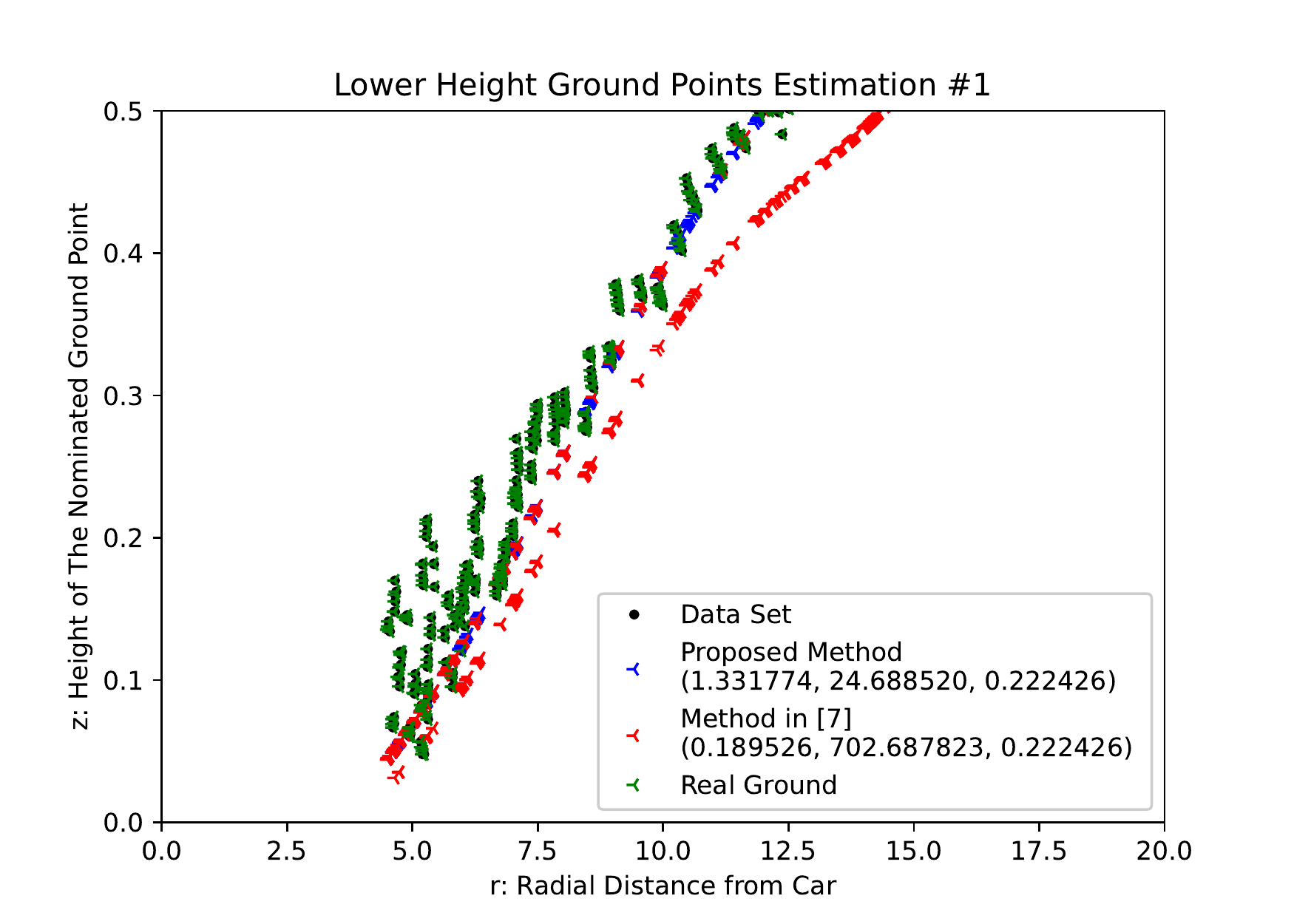}}
\end{minipage}%
\begin{minipage}{.5\linewidth}
\centering
\subfloat[]{\label{main:b}\includegraphics[trim={1cm 0 1.5cm 1cm},clip,scale=.275]{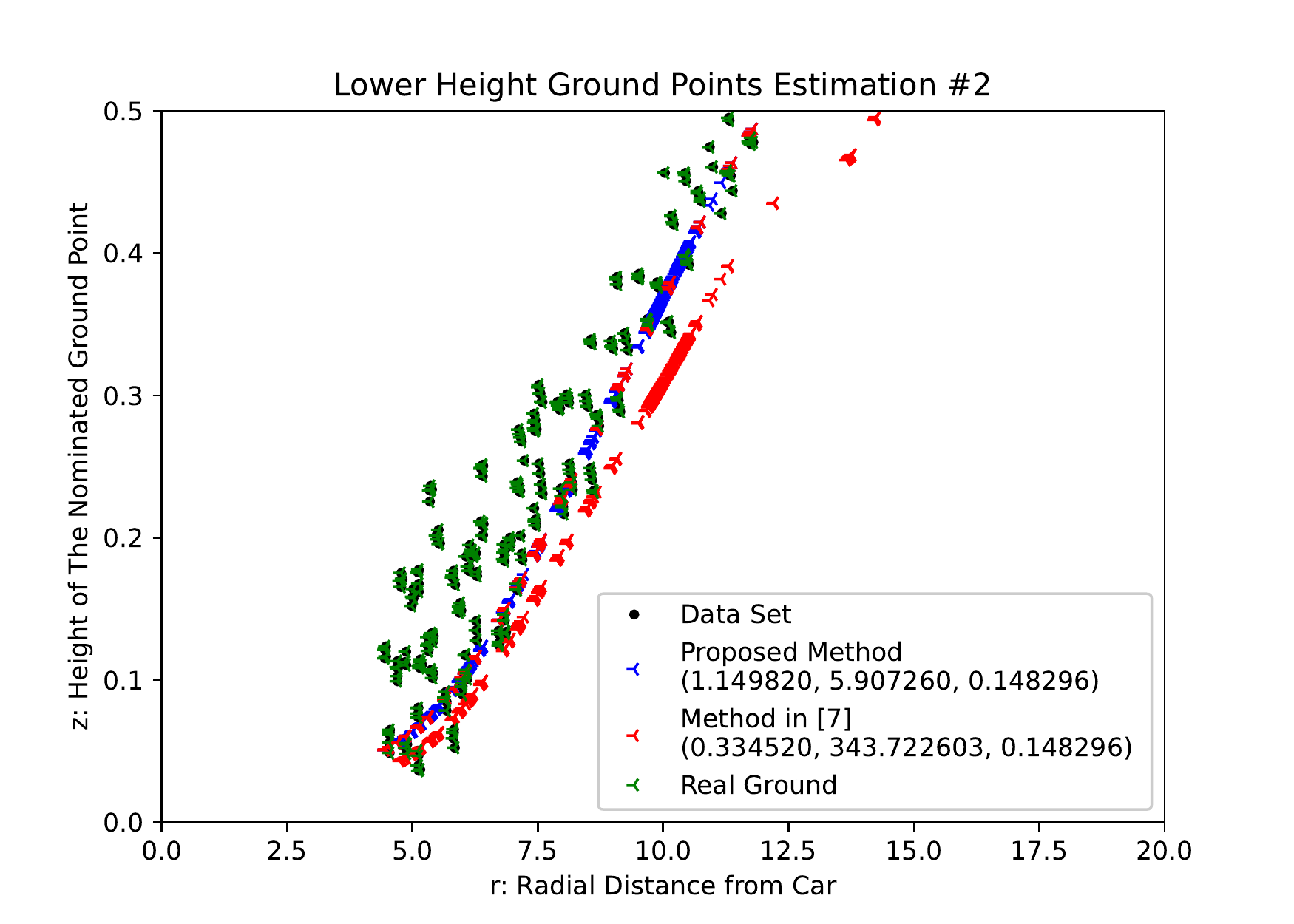}}
\end{minipage}\par \vspace{-1mm}
\centering
\subfloat[]{\label{main:c}\includegraphics[trim={1cm 0 0.8cm 1cm},clip,scale=.3]{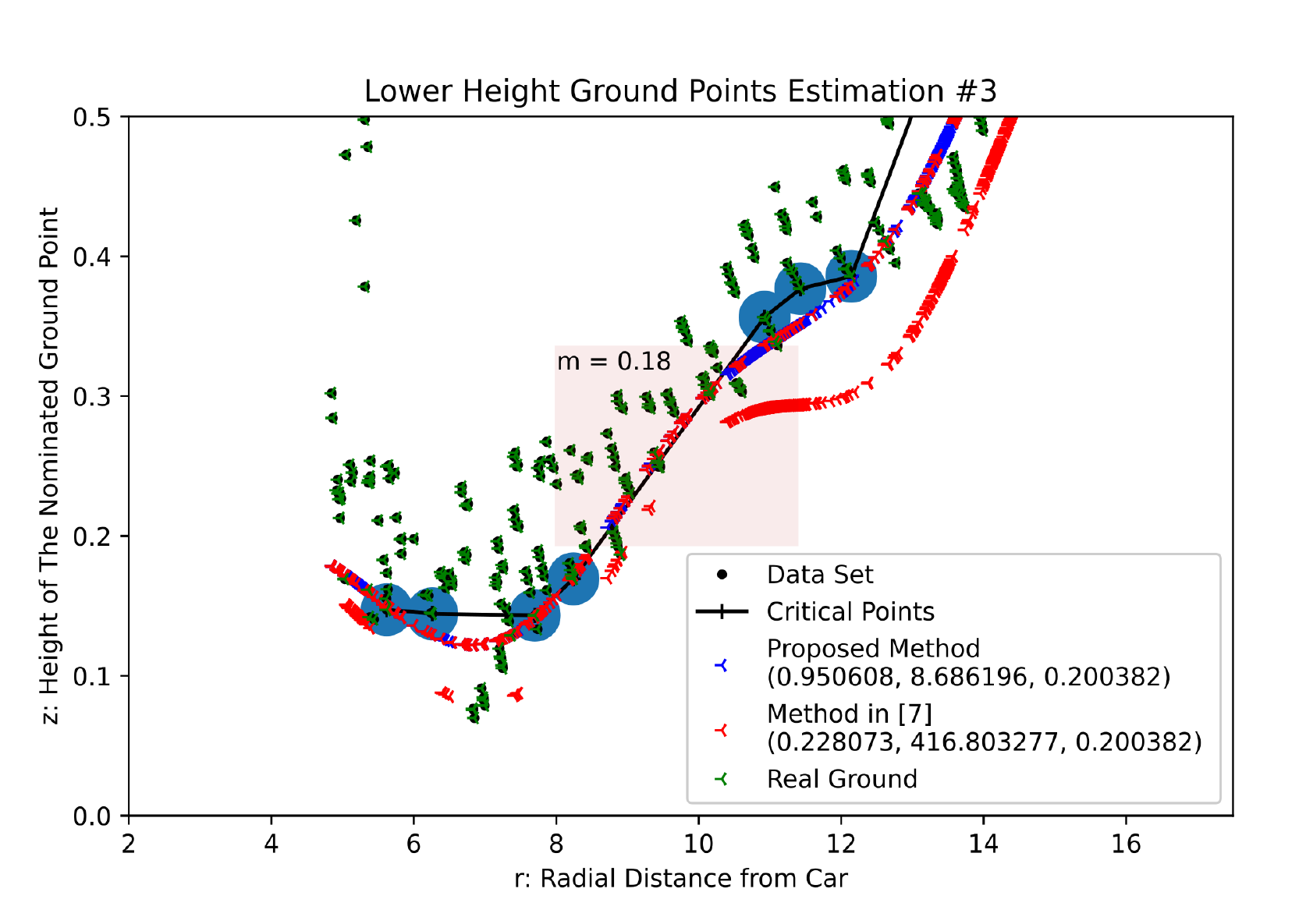}}
\caption{Implementation result of the ground segmentation method on
three different segments of data with sloped terrains. Green points
are real ground points labeled in~\cite{Lai2009}. Red dots are the
point estimation of ground height in input test locations $r_\star$
while blue points are the results of the method proposed in this
paper. In \textbf{c)} Bigger points are the critical points detected
in that segments.} \label{figure:lowheight}
\end{figure}

\paragraph{Ground Height Estimation}
Predictive distribution of ground height can be addressed after
obtaining length-scales on locations of ground candidate points.
Predictive distribution $P_m(z_*|r_*,\mathcal{R}_m)$ enables the
prediction of $z^*$ value at arbitrary locations $r^*$ at each point
cloud frame:
\begin{equation}
\mu_{\mathcal{GP}} = \bar{z} = K(r^*,r)^T\bigg[K(r,r)+
\sigma_n^2I\bigg]^{-1}z,
\end{equation}
in which, K is the covariance matrix for $\mathcal{GP}_z$ and
$\sigma_n^2$ is the measurement noise.
where, $\mathcal{L} \in \mathbb{R}$ is length scale for every data
point.
\begin{figure*}[ht]
    \centering
    \includegraphics[trim={1cm 0 1.5cm 1cm},clip,scale= 0.4]{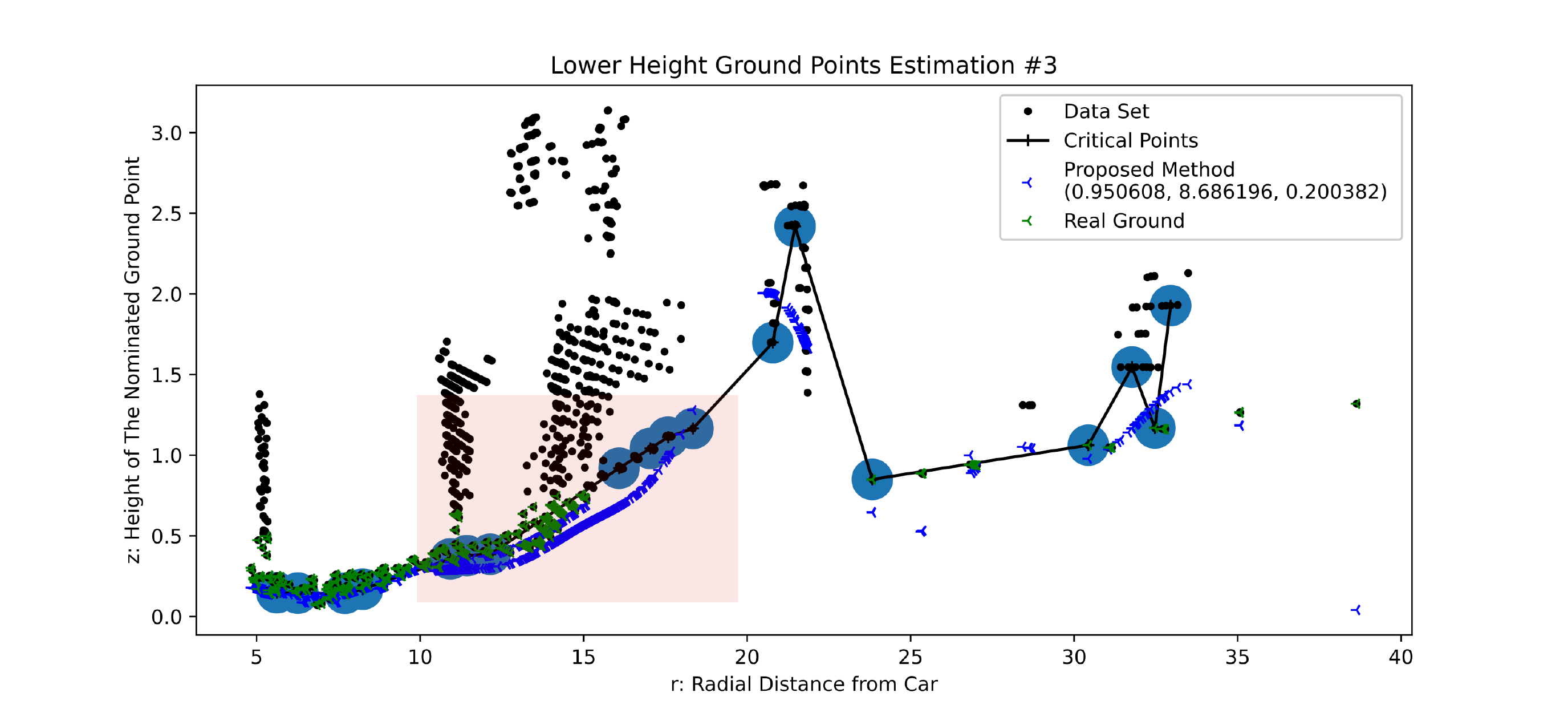}
    \caption{Estimation result of the proposed method. The big blue points are the critical points. The highlighted rectangle shows an area that ground points are placed near an obstacle. In this segment, the proposed method truly identified all 386 ground points but 9 of them giving an overall Success Rate (SR) of $97.67\%$.}
    \label{figure:overal}
\end{figure*}
\subsection{Learning Hyper--Parameters}
The behavior of the proposed model and its ability to adjust itself
to the physical realities of the environment is directly affected by
the values of hyper--parameters. Often in real-world applications,
there is no exact prior knowledge about hyper--parameters' values
and they must be obtained from data. Hyper--parameters $\theta =
\{\sigma_f,\alpha, \sigma_n\}$ are mutually independent variables to
allow the gradient-based optimization to hold its credibility. The
\textit{maximum a posteriori} argument is used to find the hyper--parameters
for ground segmentation. The hyper--parameters that maximize the
evidence of observing $z$ given $r$ or $log P(z|r,\mathcal{L})$.
Gradient-based optimization methods are used to find the
corresponding solutions.

\subsection{Whole--Frame Objective Function} If optimization process
is to be effective enough, it shall take all the segments into
account to form an objective function. This would be an example of
multi-task regression problem. The objective function in multi-task
problems is equal to the sum of all objective functions regarding to
different tasks of the problem that share the same
hyper--parameters. Therefore, we assume that all the segments share
the same hyper--parameters and establish a global view to our frame
data in order to have the results to be whole--frame inclusive.
\begin{eqnarray}
J(\theta) & = & \sum\limits_{m=1}^M \bigg(\log P(z^m|r^m,\theta)\bigg)  \\
 &  = &  -\frac{1}{2}\bigg[\sum\limits_{m=1}^M\bigg((z^m)^TA_m^{-1}z^m\bigg)+\log\bigg(\prod\limits_{m=1}^M |A_m|\bigg) \nonumber \\
 & + &\log(2\pi)\bigg(\sum\limits_{m=1}^M(n_m)\bigg)\bigg]\nonumber
\end{eqnarray}
where $A_m:=K_m(r,r)+\sigma_n^2I$ is the
corresponding covariance function of Gaussian processes in each
segment. Therefore, in every segment the ground candidate set with
assigned covariance kernels form the segment-wise objective
function. Sum of all segment-wised objective functions will form the
whole-frame objective function. For the whole-frame objective
function to hold its credibility, the hyper-parameters of the
regression task must be shared among all segments.

\section{Simulation and Results}
The method is fully developed in C++. All modules and algorithms are
implemented and developed from scratch using C++ in Qt Creator.
Furthermore, 3D point clouds are handled by the use of Point Cloud
Library (PCL). A Gaussian process regression ($\mathcal{GP}$) library
is also developed in C++ based on the math provided in
\cite{C.E.Rasmussen2006}. The block diagram of the implementation of
the proposed method is presented in Figure~\ref{figure:diagram}.
Data set in~\cite{Lai2009} is utilized to evaluate the results. To
present the efficiency of the method for sloped terrains even in low
height ground sections, segments are chosen that have slopes with
more than 10 degrees or 18\% inclination. The low height ground
estimation result is presented in Figure~\ref{figure:lowheight}. It
is depicted that the proposed method is giving better estimation
results for the sloped terrains. The average success rate of the proposed method is $93.5\%$ for sloped areas while the rate for flat areas are equal to $96.7$. The reported value from other methods is depicted in Table~\ref{tabletime}. The best optimized kernel
parameters $\theta = (\sigma_f, a, \sigma_n)$ is reported as well
for each segment.

The lower height simulation results for the sloped sections of the data
are presented as in many available methods in the literature the evaluation
results are presented for the height below $0.3m$. In Figure~\ref{main:a} and~\ref{main:b} 
The simulation results of our proposed method are presented along with the simulation results of the method presented in~\cite{Chen2014} for the ground points up to $0.5m$ of height. The result proves the effectiveness of the proposed method to detect ground points even in sloped terrains. In Figure~\ref{main:c} results of the method are shown for a piece of data that both includes flat and sloped ground. Detected critical points are also depicted in the figure. 
The slope for $3_{rd}$ the line-segment is near zero while the slope for $4_{th}$ and $5_{th}$ line-segment is more than $18\%$ (11 degrees). But the test points belonging to the $5_{th}$ line-segment (Highlighted by the rectangle) are classified as ground points. This is because the point density above this line-segment is not high, therefore the line-segment must be of a sloped ground, not an obstacle. 
\begin{table*}[t]
\centering
 \begin{tabular}{||l c c ||} 
 \hline
 Method & Average Processing Time per Frame (ms) & Average SR(flat/sloped)\\ [0.5ex] 
 \hline\hline
 Fast Ground Segmentation Method based on Jump-Convolution Network \cite{MinhChu2019b}  & 4.7  & 94.61/91.25\\ 
 Loopy Belief Propagation Based Ground Segmentation \cite{Zhang2015} & 1000 & 97.19/NR\\
 Voxel-based Ground Segmentation \cite{Cho2014} & 19.31 & NR\\
 Enhanced Ground Segmentation in Human-centric Robots \cite{MinhChu2019} & 6.9 & 94.71/91.70\\
 Fast segmentation of 3D point clouds for ground vehicles \cite{Himmelsbach2010} & 100 & NR\\
  GP-Based Real-Time Ground Segmentation for ALVs \cite{Chen2014} & 
  $\thickapprox$ 200 & 97.67 / NR\\
 Proposed Method & 72.91 & 96.7/93.5\\[1ex] 
 \hline
 \end{tabular}
     \caption{Average processing time and Success Rate (SR) of different ground segmentation methods.}
       \label{tabletime}
\end{table*}
Furthermore, as depicted in Figure~\ref{figure:overal} the results show that the proposed algorithm is effective on the whole data without any data truncation or other pre-processings leading to data omission, the
advantage of which is the detection of the ground points which have
greater heights or distance from the car. For example, most of the ground segmentation methods in the literature neglect all the points higher than $30cm$.
A line with $30cm$ height is depicted in Figure~\ref{main:c} and Figure~\ref{figure:overal}. The method may neglect more than half of the ground points just by neglecting the heights greater than $0.3m$ Although it is depicted that the ground points higher than $30cm$ are well detected
by the proposed method. On the other hand, the ground point might be
found in the far distance from the car as depicted in
Figure~\ref{figure:overal}. 

Furthermore, In the colored rectangle at the distance $10$ to $20m$ in Figure~\ref{figure:overal} the ground data is juxtaposed to an object's points while both having the same range of height and gradient values. The algorithm finds it hard to maintain a
continuous estimation of the ground as it is alerted by the value of $\tilde{\rho}$ that an obstacle might be placed here. The reason is that in this
line-segment the density-like parameter $\tilde{\rho}$ is signaling
the GP that a high number of points are available at the top of the
ground candidate points, Therefore the estimation tries to keep its
results near to the real ground points at the beginning but suddenly
keeps its distance to the obstacle points. 

Table~\ref{tabletime} compares the processing time of different ground segmentation methods. Each frame of the data set presented by \cite{Lai2009} contains approximately 70,000 three-dimensional points. The Velodyne LiDAR sensor captures data at 10 $fps$. The reported average processing time of the proposed method is based on the implementation of an Intel Core i7-7700HQ CPU @ 2.80GHz. The proposed method is able to operate at 14 $fps$ which is about 1.5 times faster than the capture rate of the LiDAR sensors, making it suitable for real-time applications.

\section{Conclusions}
A Physically-motivated ground segmentation algorithm is proposed for sloped terrains. The 3D LiDAR data is categorized based on a radial grid map to construct a segment-wise representation of 2D points. A line-segment dividing strategy is implemented on the 2D points of
each segment based on the detection of critical points. The
parameters regarding the line in each segment are obtained by a
two-dimensional line regression algorithm that assumes non-equal
error in both directions of the coordinate system. On the other
hand, distance-related line fitting is utilized for the algorithm to
be more effective in larger slopes. Furthermore, a density-like
parameter is introduced to handle situations in which ground
candidate points are just the minimum height points of on-road
obstacles. Then based on these parameters a non-stationary
covariance function is constructed and is fed to Gaussian process
regression to obtain ground estimation. The method is tested on the
labeled data set provided by~\cite{Lai2009}. The effectiveness of
the proposed method is verified by the results of the test which
outperforms similar methods in the sloped area and in the overall
determination of ground points even when no data truncation is
performed.

\bibliographystyle{./bibliography/IEEEtran}
\bibliography{./bibliography/ALVGS}

\end{document}